%% file: vwen.tex
\documentclass[11pt]{article}

\usepackage[preprint]{acl}

\usepackage{times}
\usepackage{latexsym}
\usepackage[T5,T1]{fontenc}            
\usepackage[utf8]{inputenc}
\usepackage[vietnamese,english]{babel}  
\usepackage{microtype}
\usepackage{inconsolata}
\usepackage{graphicx}
\usepackage{booktabs}
\usepackage{multirow}
\usepackage{amsmath}
\usepackage{xcolor}
\usepackage{pifont}

\graphicspath{{vwen/figures/}}

\newcommand{\vi}[1]{{\fontencoding{T5}\selectfont\foreignlanguage{vietnamese}{#1}}} 
\newcommand{\cmark}{\ding{51}}
\newcommand{\xmark}{\ding{55}}
\definecolor{hlcolor}{RGB}{253,227,167}
\newcommand{\nw}[1]{{\setlength{\fboxsep}{0.6pt}\colorbox{hlcolor}{#1}}} 
\newcommand{\best}[1]{\textbf{#1}}
\newcommand{\Dv}{\ensuremath{D}}
\newcommand{\at}[1]{\,@#1}

\title{VWEN: A Curated Small Language Model for Vietnamese Paraphrasing}

\author{Sang Quang Nguyen \\
\textit{University of Information Technology}\\
\textit{Vietnam National University}\\
\texttt{sangnq.19@grad.uit.edu.vn}}

\begin{document}
\maketitle

\begin{abstract}
We present VWEN, a 0.8B-parameter model for Vietnamese paraphrasing at a requested level of change and for five related
tasks: generating several paraphrases, simplification, gist, register transfer and paraphrase judgement. VWEN is
Qwen3.5-0.8B fine-tuned on about 7k samples chosen by a four-stage pipeline: rules and deduplication, a bidirectional
entailment filter, a divergence score that assigns low and high paraphrase levels, and selection for coverage and
instruction-following difficulty. We compare it with paraphrase-only recipes from the same pipeline and with training on
the ViSP corpus under matched and larger step budgets. On high-divergence paraphrasing, VWEN reaches a success rate
(requested level reached and meaning preserved) of 0.61 against 0.46 for the ViSP model at matched compute, a difference
that holds on inputs resembling ViSP. Given four times more training examples, the ViSP model closes the gap on such inputs
but not on questions, which ViSP lacks and which it often turns into statements. Across all comparisons the entailment
filter is the most effective stage. Training on paraphrases alone erases abilities of the base model that the
multi-task data preserves.
\end{abstract}

\section{Introduction}
\label{sec:intro}

Vietnamese paraphrasing has large-scale data: ViSP \citep{nguyen-nguyen-2025-visp} collects 1.2M pairs when reversed
and cross pairs are counted, and its expanded training split holds 406,079 pairs. Work on instruction tuning suggests that
the quality and diversity of a small set can matter more than size \citep{zhou2023lima,chen2024alpagasus,li-etal-2024-quantity,liu2024deita},
but these results come from general-purpose chat models, mostly in English, and
measure overall preference rather than a specific behaviour that can be checked.

Paraphrasing at a requested level of change is such a behaviour: given an input, a model should change it a little or a
lot on request and keep its meaning, and both properties can be measured. A practical Vietnamese writing assistant also
needs related rewrites (simplify, summarise, shift register) and a way to tell whether two sentences mean the same.
Models under 1B parameters can run on a laptop CPU and suit this use, but they easily lose existing abilities when
fine-tuned on a single task.

We build such a model from a few thousand selected samples and ask three questions. \textbf{Q1}: at matched compute,
does the selected data do better than training on ViSP, and on which inputs? \textbf{Q2}: which pipeline stage contributes?
\textbf{Q3}: how many training examples does each recipe need?

\paragraph{Contributions.} (1) VWEN, a 0.8B Vietnamese rewriting model, and the four-stage pipeline that selects its
training data, with rejection counts at every stage. (2) An evaluation suite that requests both levels for the same input
and scores level separation, meaning preservation and their conjunction, plus a meaning $\times$ divergence grid and targeted
perturbations for paraphrase judgement. (3) A five-recipe comparison with paired significance tests, a source-matched
random baseline, and a split by input type that exposes a distribution gap between ViSP and question paraphrasing. (4) An
error analysis showing that bidirectional entailment accepts outputs that answer a question instead of rephrasing it.

The model is available at \url{https://huggingface.co/ngwgsang/vwen-r4}.

\section{Related Work}
\label{sec:related}

\paragraph{Vietnamese paraphrase resources.} ViSP \citep{nguyen-nguyen-2025-visp} is a 1.2M-pair sentence paraphrase
benchmark with seq2seq and LLM baselines; ViQP \citep{nguyen2023viqp} pairs human-written question paraphrases; VietQuill
\citep{nguyen2026vietquill} contains sentence and question paraphrases synthesised in the quality-controlled style of
QCPG \citep{bandel-etal-2022-quality}; ViePAWS \citep{nguyen2026viepaws} adapts the adversarial pairs of PAWS
\citep{zhang-etal-2019-paws} to Vietnamese; ViANLI \citep{huynh2024vianli} is an adversarial Vietnamese NLI set. We reuse
all of them as candidate pools and evaluation sources.

\paragraph{Data selection for instruction tuning.} LIMA aligns a model with about 1k curated examples
\citep{zhou2023lima}; AlpaGasus filters with an LLM grader \citep{chen2024alpagasus}; IFD scores how much the instruction
helps predict the answer \citep{li-etal-2024-quantity}; DEITA combines complexity, quality and diversity
\citep{liu2024deita}. We use IFD and an LLM grader as pipeline stages and add two stages specific to paraphrasing:
bidirectional entailment and a divergence score.

\paragraph{Controllable paraphrasing.} Prior work controls the target syntax \citep{iyyer-etal-2018-adversarial} or a
vector of semantic, syntactic and lexical quality \citep{bandel-etal-2022-quality}. We control the level through
natural-language instructions and measure divergence with dependency tree edit distance \citep{pawlik2016tree} combined
with syllable edit distance.

\paragraph{Evaluating paraphrases.} \citet{shen-etal-2022-evaluation} show that common paraphrase metrics agree poorly
with human judgements and that lexical divergence must be modelled explicitly; we score meaning and divergence separately
and combine them. Entailment-based meaning checks are a common proxy; Section~\ref{sec:errors} documents a failure of
this proxy on question inputs.

\section{Tasks, Notation and Data}
\label{sec:data}

\begin{table}[t]
\centering\small
\begin{tabular}{@{}lp{0.47\columnwidth}r@{}}
\toprule
Task & Description (output) & Quota \\
\midrule
Para-L & Low-divergence paraphrase: small edits, structure kept (one sentence) & 1,350 \\
Para-H & High-divergence paraphrase: new wording and structure (one sentence) & 1,350 \\
Multi-Para & $k$ distinct paraphrases (list) & 700 \\
Simplify & Simplify for a general reader, keep all content (one sentence) & 900 \\
Gist & Main idea; minor details may be dropped (short sentence) & 600 \\
Style & Register transfer, formal $\leftrightarrow$ informal (one sentence) & 900 \\
Equiv & Do the two sentences mean the same? (yes / no) & 1,500 \\
\bottomrule
\end{tabular}
\caption{Tasks and target quotas at the 8k budget. A general-instruction task (800) planned for the budget had no
source, so the multi-task recipe has 7,087 rather than $\sim$8,000 samples.}
\label{tab:tasks}
\end{table}

\begin{table}[t]
\centering\small
\begin{tabular}{@{}lp{0.68\columnwidth}@{}}
\toprule
Symbol & Meaning \\
\midrule
$\Dv(x,y)$ & Divergence of output $y$ from input $x$: mean percentile rank of normalised dependency tree edit distance and syllable edit distance, in $[0,1]$ \\
MP & Meaning preservation: share of outputs entailed in both directions (entailment $\geq 0.80$) by an NLI model not used for data filtering; Gist uses forward entailment only \\
band(L), band(H) & $\Dv \leq 0.205$ and $\Dv \geq 0.435$, the thresholds that define the two levels in the training data \\
Success & Share of outputs in the requested band \emph{and} preserving meaning; the main paraphrase metric \\
Level gap $\Delta$ & Mean of $\Dv(x,y_H) - \Dv(x,y_L)$ when the same input $x$ is requested at both levels \\
Order & Share of inputs with $\Dv(x,y_H) > \Dv(x,y_L)$ \\
LLM rating & LLM grader score in $[0,1]$ combining meaning, fluency and instruction following, on 100 outputs per task \\
R1--R5 & Data recipes (Table~\ref{tab:recipes}); ``R5\at{1.5k}'' and ``R5\at{4k}'' denote the step budget \\
\bottomrule
\end{tabular}
\caption{Notation.}
\label{tab:notation}
\end{table}

Table~\ref{tab:tasks} lists the tasks and Table~\ref{tab:notation} the notation.

\paragraph{Sources.} (i) Existing datasets: ViSP (its 98k original pairs enter the candidate pool; the full 406k expanded
training pairs form recipe R5), ViQP, VietQuill, ViePAWS and ViANLI. (ii) A multi-domain source-sentence pool
(administrative, education, health, news, literature, conversation, social media), split by source document. (iii)
Outputs distilled from mid-sized open LLMs (Gemma 4 26B-A4B, \citealp{gemma2026gemma4}; Qwen3-30B-A3B,
\citealp{yang2025qwen3}) for Simplify, Gist, Style, Multi-Para and part of the paraphrase pool; smaller teachers were tried
first and compared on shared batches. (iv) Rule-based targeted perturbations (negation, numbers and dates, entities,
antonyms, causality) as negatives for Equiv, rewritten for fluency and label-checked by an LLM. (v) 634 seed samples
drafted by an LLM and reviewed by the author.

\paragraph{Leakage control.} Original splits are kept (validation $\rightarrow$ dev). A sentence that appears in test is
removed from dev and train, and one in dev is removed from train, across all datasets, including near duplicates proposed
by embedding similarity and confirmed by syllable-sequence distance. Dev and test were frozen and hashed before any
training run.

\paragraph{Evaluation sets.} Dev and test each hold 250 samples per task, balanced over control setting, source dataset,
domain and length. ViSP is used for training only, so the paraphrase inputs of dev and test come from ViQP and VietQuill
(44\% on test) and from distilled pairs (56\%). \textbf{56\% of the test paraphrase inputs are questions} (55\% on dev),
against 0.2\% of ViSP training inputs; we report results separately for the two input types (Section~\ref{sec:q1}). Equiv
has two further sets: perturbations (40 pairs per kind and per ViePAWS label) and a grid (50 pairs per cell of meaning
\{preserved, drifted\} $\times$ \Dv\ \{low, mid, high\}, drawn before the meaning filter). No sentence pair is shared
between sets.

\section{Selection Pipeline}
\label{sec:pipeline}

\begin{table*}[t]
\centering\small
\begin{tabular}{@{}lrrp{0.5\textwidth}@{}}
\toprule
Stage & In & Kept & Main rejection reasons \\
\midrule
0. Candidates & -- & 341,294 & \\
1. Rules, deduplication, leakage & 341,294 & 296,081 & near duplicate within task 30,474; leakage to test 4,330 and dev 1,656; copied input 3,378; length ratio 2,093 \\
2. Meaning (bidirectional NLI) & 296,081 & 205,892 & forward entailment 57,116; backward entailment 31,307; BERTScore floor 1,317 \\
3. Divergence and level assignment & 206,371 & 162,517 & middle band 43,854 (29,546 Para-L and 104,713 Para-H kept) \\
4. Coverage and IFD selection & 162,517 & 7,209 & selection to quota; 7,087 after the LLM quality grader \\
\bottomrule
\end{tabular}
\caption{Pipeline on the training split. Stage 3 assigns levels to unlevelled paraphrase pairs and drops the middle band.
Per-task counts are in Appendix~\ref{app:counts}.}
\label{tab:pipeline}
\end{table*}

Table~\ref{tab:pipeline} summarises the four stages.

\paragraph{Stage 1.} Rules remove non-Latin script, unbalanced brackets, abnormal length ratios and outputs that copy the
input. Near duplicates are proposed by multilingual-e5-small \citep{wang2024multilingual} cosine similarity and confirmed
by syllable-sequence distance, because e5 scores genuine paraphrases at 0.95--0.99 and cannot separate them from
duplicates on its own.

\paragraph{Stage 2.} Entailment from an mDeBERTa-v3 model \citep{he2023debertav3} fine-tuned on XNLI
\citep{conneau-etal-2018-xnli} and multilingual NLI data \citep{laurer2024less} must reach 0.80 in both directions (Gist:
forward only). BERTScore \citep{zhang2020bertscore} with xlm-roberta-large \citep{conneau-etal-2020-unsupervised} is kept
only as a garbage floor ($\geq 0.30$): it rewards lexical overlap, and a 0.8 threshold removes every Para-H pair.
Thresholds were calibrated on labelled pairs. Pairs that drift in both directions become negative candidates for Equiv,
used only after an LLM grader confirms them, since NLI mislabels about half of them.

\paragraph{Stage 3.} \Dv\ combines dependency tree edit distance (APTED, \citealp{pawlik2016tree}, over Stanza
\citep{qi-etal-2020-stanza} UPOS and relation labels, normalised by tree size) and syllable Levenshtein distance, each
converted to a percentile against a reference sample. Pairs with $\Dv \leq 0.205$ become Para-L, $\Dv \geq 0.435$
Para-H, and the middle band is dropped so that the two levels are well separated.

\paragraph{Stage 4.} Source sentences are clustered (k-means, 200 clusters, e5 embeddings). Within each task, samples are
ordered round-robin over control setting, domain, length bucket and cluster, so every prefix is balanced and the 1k, 2k,
4k and 8k sets are nested. From a pool of three times the quota, samples with IFD $\geq 1$
\citep[the instruction does not help predict the answer;][]{li-etal-2024-quantity} are dropped and mid-range IFD is
preferred within each cluster. All 200 clusters are represented. An LLM grader finally removes generation samples rated
below threshold on meaning, fluency or instruction following.

\section{Experimental Setup}
\label{sec:setup}

\begin{table*}[t]
\centering\small
\scalebox{0.94}{
\begin{tabular}{@{}lp{0.52\textwidth}rr@{}}
\toprule
Recipe & Data & Samples & Paraphrase samples \\
\midrule
R1 Random & random paraphrase pairs after stage 1 (before the meaning filter), levelled like R5 & 8,000 & 8,000 \\
R2 +Filter & + meaning filter and level assignment (stages 2--3) & 7,793 & 7,793 \\
R3 +Coverage & + coverage and IFD selection (stage 4) & 7,854 & 7,854 \\
R4 Multi-task (VWEN) & + the other five tasks (full pipeline) & 7,087 & 2,615 \\
R5 ViSP & ViSP training pool (406,079 pairs), levelled by \Dv\ at the midpoint of the two bands; 1,500 or 4,000 steps & 24k / 56k seen & all \\
\bottomrule
\end{tabular}
}
\caption{Data recipes at the 8k budget (actual counts after the quality grader).}
\label{tab:recipes}
\end{table*}

Table~\ref{tab:recipes} defines the five recipes. R1 is a \textbf{source-matched control}: it is drawn from the same
candidate pool as R2--R4 (56\% distilled pairs, 44\% existing datasets including ViSP originals), before any meaning
filtering, and is levelled the same way as R5.

\paragraph{Training.} Qwen3.5-0.8B \citep{qwen2026qwen35}, full fine-tuning on one 16GB T4: float32 weights with fp16
mixed precision, 8-bit AdamW, gradient checkpointing, learning rate 1e-5 with cosine decay and 5\% warmup, effective batch
16, at most three epochs, loss on the answer only, reasoning mode disabled. Validation loss is computed twice per epoch on
50 dev samples per trained task; training stops after two evaluations without improvement and the best checkpoint is
restored. R5 runs a fixed step budget without early stopping (its long cosine schedule otherwise triggers spurious stops)
and also restores its best checkpoint.

\paragraph{Compute.} 1,500 steps is the three-epoch cap at the 8k budget, so R5\at{1.5k} matches the maximum compute of
R1--R4. Early stopping makes R1--R4 cheaper in practice: the R4 checkpoint used at 8k saw 14,144 examples, R5\at{1.5k}
saw 24,000 and R5\at{4k} saw 56,000. We report \emph{examples seen by the selected checkpoint} (best step $\times$ batch
size) alongside nominal set size (Table~\ref{tab:seen}).

\paragraph{Inference and scoring.} Greedy decoding. Each Para-L and Para-H input is also requested at the other level,
giving both outputs for the same input. MP uses an XLM-R large model fine-tuned on XNLI
\citep{conneau-etal-2018-xnli,conneau-etal-2020-unsupervised}, deliberately different from the filtering NLI model.
Simplify, Gist and Style are rated by an LLM grader on 100 outputs per task. Every test input is also run with a held-out
instruction phrasing that never appears in training (Section~\ref{sec:heldout}).

\paragraph{Statistics.} R4 at 8k and R5\at{1.5k} have three seeds (42, 43, 44); all other runs have one, for GPU budget
reasons. Differences between recipes are tested with a \textbf{paired bootstrap over test inputs} (10,000 resamples):
for each input we take the per-run score, averaged over seeds when there are three, and resample inputs. This accounts
for evaluation-sample variance; seed variance enters only through the averaging. We report 95\% intervals and do not
correct for multiple comparisons.

\section{Results}
\label{sec:results}

\subsection{Q1: Selected Data against ViSP}
\label{sec:q1}

\begin{table*}[t]
\centering\small
\scalebox{0.88}{
\begin{tabular}{@{}llrrrrrrr@{}}
\toprule
Task & Metric & Base & R1 & R2 & R3 & R4 & R5\at{1.5k} & R5\at{4k} \\
\midrule
Para & Level gap $\Delta$ & 0.207 & 0.296 & 0.402 & \best{0.416} & 0.342 $\pm$ 0.009 & 0.297 $\pm$ 0.004 & 0.327 \\
Para & Order & 0.796 & 0.834 & 0.900 & \best{0.940} & 0.843 & 0.786 & 0.816 \\
\midrule
Para-H & Success & 0.084 & 0.540 & \best{0.708} & 0.644 & 0.609 $\pm$ 0.017 & 0.459 $\pm$ 0.022 & 0.528 \\
Para-H & MP & 0.688 & 0.856 & 0.892 & \best{0.920} & \best{0.920} $\pm$ 0.014 & 0.865 $\pm$ 0.027 & 0.876 \\
Para-H & \Dv & 0.264 & 0.516 & \best{0.623} & 0.564 & 0.553 & 0.472 & 0.487 \\
\midrule
Para-L & Success & 0.904 & 0.664 & 0.716 & \best{0.808} & 0.728 $\pm$ 0.004 & 0.749 $\pm$ 0.033 & 0.756 \\
Para-L & MP & 0.936 & 0.904 & 0.900 & \best{0.956} & 0.940 $\pm$ 0.017 & 0.921 $\pm$ 0.029 & 0.944 \\
\midrule
Multi-Para & Count accuracy & 0.52 & 0 & 0 & 0 & \best{1.00} & 0 & 0 \\
Simplify & LLM rating & 0.713 & 0.725 & 0.730 & \best{0.838} & 0.834 & 0.814 & 0.797 \\
Gist & LLM rating & 0.73 & 0.46 & 0.54 & 0.48 & \best{0.82} & 0.51 & 0.51 \\
Style & LLM rating & 0.527 & 0.445 & 0.515 & 0.537 & \best{0.688} & 0.526 & 0.565 \\
Equiv & Accuracy (invalid) & 0.748 (0) & 0.16 (0.71) & 0.14 (0.73) & 0.16 (0.74) & \best{0.929} (0) & 0.163 (0.73) & 0.172 (0.72) \\
\bottomrule
\end{tabular}
}
\caption{Test results, canonical instructions (250 inputs per task; $\Delta$ and Order over 500 paired requests). R4 and
R5\at{1.5k}: mean $\pm$ sd over three seeds; other recipes: one seed. Bold marks the best fine-tuned value.}
\label{tab:main}
\end{table*}

\begin{table*}[t]
\centering\small
\begin{tabular}{@{}llrrr@{}}
\toprule
Comparison & Metric & All inputs & Non-questions & Questions \\
\midrule
R4 $-$ R5\at{1.5k} & Para-H success & \best{+0.151} [0.105, 0.197] & \best{+0.156} [0.082, 0.231] & \best{+0.147} [0.088, 0.206] \\
 & Para-H MP & \best{+0.055} [0.024, 0.085] & \best{+0.061} [0.020, 0.102] & \best{+0.050} [0.009, 0.092] \\
 & Level gap $\Delta$ & \best{+0.045} [0.024, 0.066] & \best{+0.067} [0.034, 0.100] & +0.028 [$-$0.001, 0.055] \\
 & Para-L success & $-$0.021 [$-$0.073, 0.029] & +0.014 [$-$0.052, 0.080] & $-$0.054 [$-$0.132, 0.023] \\
\midrule
R4 $-$ R5\at{4k} & Para-H success & \best{+0.081} [0.011, 0.151] & +0.048 [$-$0.061, 0.156] & \best{+0.103} [0.018, 0.189] \\
 & Para-H MP & \best{+0.044} [0.001, 0.088] & +0.007 [$-$0.041, 0.065] & \best{+0.068} [0.007, 0.132] \\
 & Level gap $\Delta$ & +0.015 [$-$0.012, 0.042] & +0.042 [$-$0.001, 0.086] & $-$0.006 [$-$0.040, 0.029] \\
 & Para-L success & $-$0.028 [$-$0.088, 0.032] & +0.058 [$-$0.030, 0.146] & \best{$-$0.109} [$-$0.191, $-$0.026] \\
\midrule
R1 $-$ R5\at{1.5k} & Para-H success & \best{+0.081} [0.016, 0.147] & $-$0.024 [$-$0.122, 0.075] & \best{+0.149} [0.061, 0.237] \\
 & Level gap $\Delta$ & $-$0.001 [$-$0.028, 0.026] & $-$0.021 [$-$0.059, 0.018] & +0.014 [$-$0.022, 0.051] \\
 & Para-L success & \best{$-$0.085} [$-$0.149, $-$0.024] & $-$0.044 [$-$0.132, 0.041] & \best{$-$0.124} [$-$0.212, $-$0.036] \\
\midrule
R2 $-$ R1 & Para-H success & \best{+0.168} [0.088, 0.244] & \best{+0.265} [0.143, 0.388] & \best{+0.105} [0.007, 0.197] \\
 & Level gap $\Delta$ & \best{+0.106} [0.073, 0.138] & \best{+0.130} [0.079, 0.182] & \best{+0.087} [0.046, 0.129] \\
\bottomrule
\end{tabular}
\caption{Paired differences on test (first recipe minus second), with 95\% paired-bootstrap intervals, on all inputs and
split by input type. Bold: interval excludes zero. $n = 250$ inputs for Para-H (152 questions), 500 requests for
$\Delta$ (281 on questions).}
\label{tab:paired}
\end{table*}

Table~\ref{tab:main} gives the test results and Table~\ref{tab:paired} the paired differences.

\paragraph{At matched compute, VWEN is better on high-divergence paraphrasing, on both input types.} Against
R5\at{1.5k}, R4 raises Para-H success by 0.15, Para-H meaning preservation by 0.06 and the level gap by 0.05, all with
intervals excluding zero, while using fewer examples (14k vs.\ 24k seen). The Para-H difference is the same on
non-questions (+0.16), the inputs that resemble ViSP, so it is not explained by the question gap alone. Para-L does not
differ.

\paragraph{With four times more examples, the ViSP model catches up on its own kind of input.} R5\at{4k} (56k examples
seen, $\sim$2.6$\times$ the training time of R4) shows no significant difference from R4 on non-questions on any
paraphrase metric. Its remaining deficit is on questions (Para-H success $-$0.10, MP $-$0.07), where it is better than R4
on Para-L success (+0.11). We therefore do not claim that selected data beats ViSP in general, only that it reaches the
same level with fewer examples and covers input types ViSP lacks.

\paragraph{The data source matters mainly for questions.} The source-matched control R1 equals R5\at{1.5k} on
non-questions on every metric, and differs only on questions, in both directions (Para-H success +0.15, Para-L success
$-$0.12). Training on a pool that contains question paraphrases helps rewrite questions boldly; it does not by itself help
on other inputs.

\paragraph{Beyond paraphrasing.} R5 never sees Simplify, Gist, Style, Multi-Para or Equiv, so R4's lead there is
expected. The informative point is that every paraphrase-only model falls \emph{below} the base model on Gist (0.46--0.54
vs.\ 0.73) and on Equiv, where it answers by rewriting the question instead of giving a label (72\% invalid outputs),
while R4 keeps and improves both (Gist 0.82, Equiv 0.929).

\begin{figure*}[t]
\centering
\includegraphics[width=0.85\textwidth]{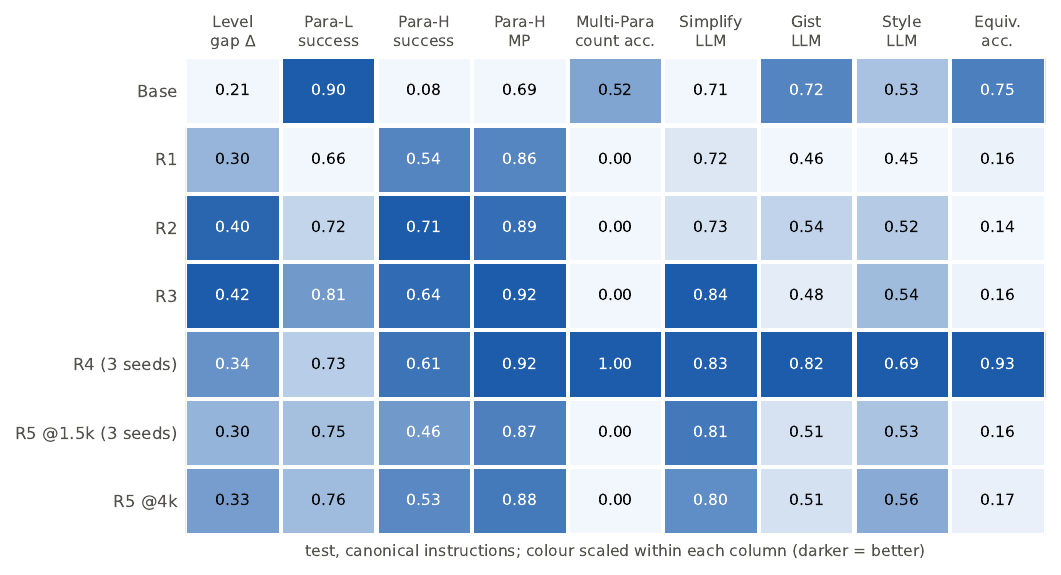}
\caption{Recipe $\times$ task on test. Colour is scaled within each column (darker is better). The base model's high
Para-L success reflects near-copies, which fall inside the low band (Section~\ref{sec:validity}).}
\label{fig:heatmap}
\end{figure*}

\subsection{Q2: Which Pipeline Stage Helps}
\label{sec:q2}

\paragraph{The meaning filter is the largest and most robust step.} From R1 to R2, Para-H success rises by 0.17 and the
level gap by 0.11, with intervals excluding zero on all inputs, on non-questions (+0.27, +0.13) and on questions (+0.11,
+0.09) (Table~\ref{tab:paired}). It is the only stage whose effect holds on every slice.

\paragraph{Coverage and IFD selection} (R2 $\rightarrow$ R3) mostly improve meaning preservation (Para-L MP 0.900
$\rightarrow$ 0.956, Para-H MP 0.892 $\rightarrow$ 0.920) and Simplify (0.730 $\rightarrow$ 0.838); these are
single-seed differences and we do not rank R2 against R3 on Para-H success, where test and dev disagree.

\begin{figure}[t]
\centering
\includegraphics[width=\columnwidth]{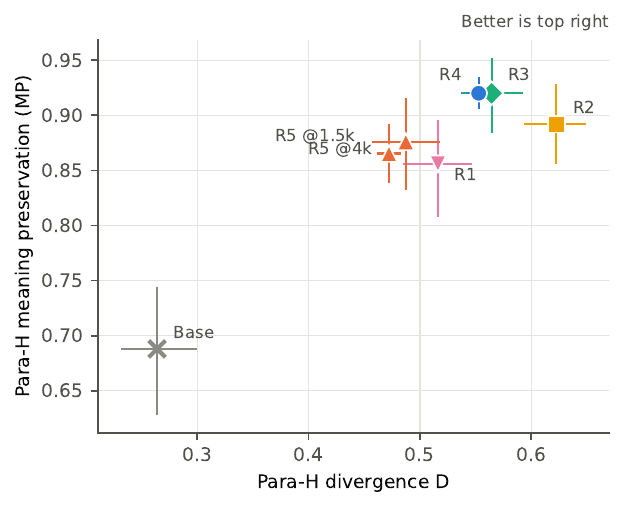}
\caption{Para-H on test: divergence \Dv\ against meaning preservation. Error bars: seed sd for R4 and R5\at{1.5k},
bootstrap intervals otherwise.}
\label{fig:meaning-diff}
\end{figure}

\begin{figure}[t]
\centering
\includegraphics[width=\columnwidth]{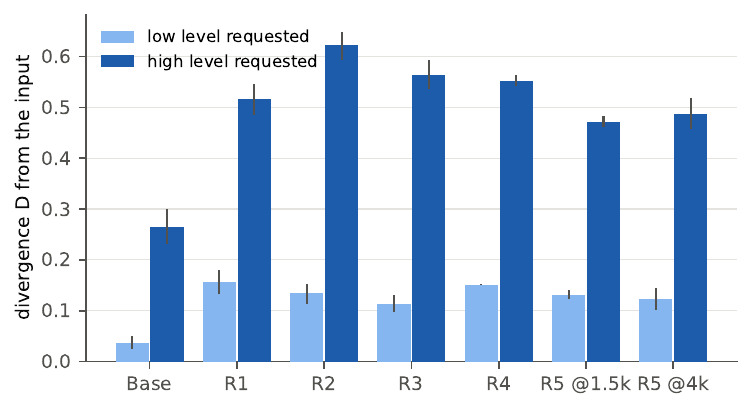}
\caption{Divergence of outputs when the low and the high level are requested for the same inputs (test).}
\label{fig:control}
\end{figure}

\paragraph{Adding tasks at a fixed budget trades level control for breadth.} At 8k, R3 beats R4 on level gap (+0.074
[0.049, 0.099]) and Para-L success (+0.080 [0.025, 0.135]) but not on Para-H success (+0.035 [$-$0.027, 0.097]) or MP:
R4 has a third of R3's paraphrase samples. Figures~\ref{fig:meaning-diff} and~\ref{fig:control} show the trade-off
between divergence and meaning and the separation between the two requested levels.

\subsection{Q3: How Many Examples Are Needed}
\label{sec:q3}

\begin{figure*}[t]
\centering
\includegraphics[width=0.75\textwidth]{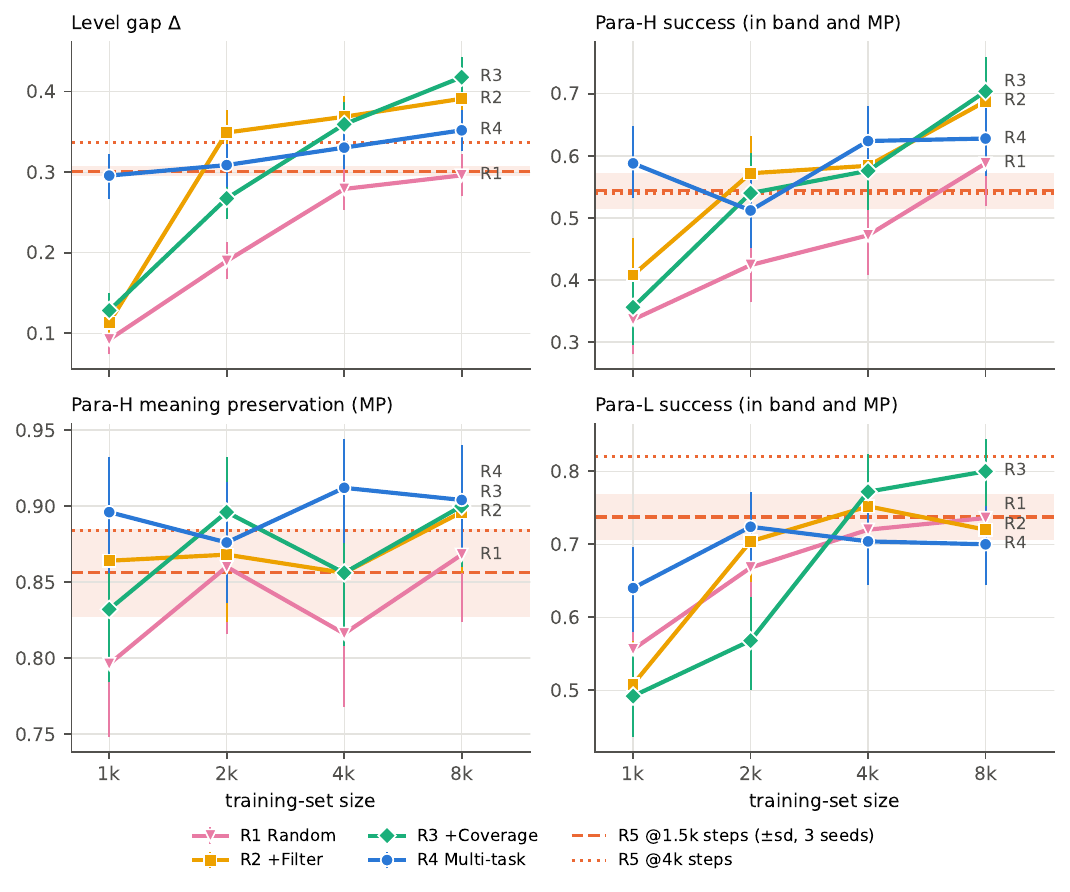}
\caption{Dev metrics by nominal training-set size (seed 42; bootstrap intervals). R5 is shown as horizontal references:
dashed @1.5k (band: $\pm$ sd over three seeds), dotted @4k.}
\label{fig:scaling}
\end{figure*}

\begin{figure*}[t]
\centering
\includegraphics[width=\textwidth]{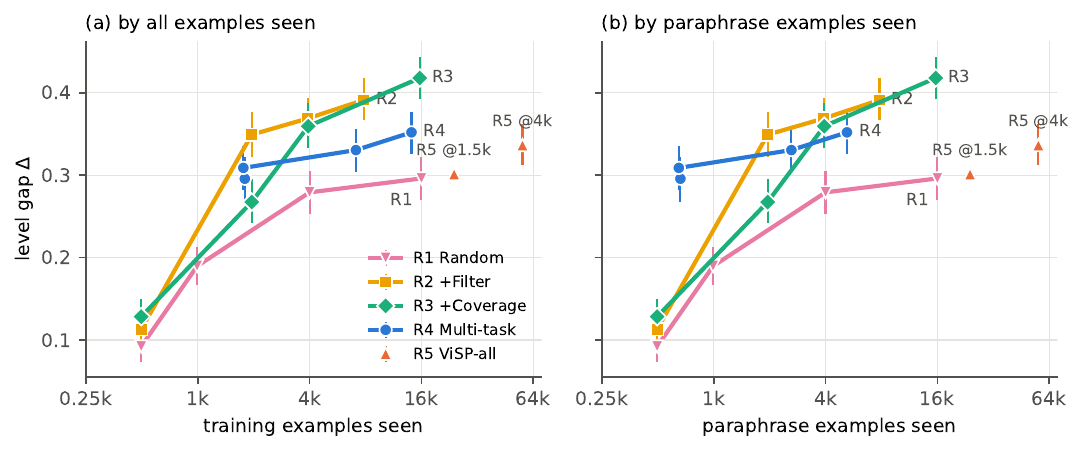}
\caption{Dev level gap against training examples (a) and paraphrase examples (b) seen by the selected checkpoint (best
step $\times$ batch size).}
\label{fig:seen}
\end{figure*}

\paragraph{Nominal size is misleading under early stopping.} By nominal size (Figure~\ref{fig:scaling}), R4 at 1k
reaches $\Delta = 0.296$ while R1--R3 at 1k reach only 0.09--0.13. But the paraphrase-only runs at 1k stopped after half
an epoch (496 examples seen) because their validation loss was flat, whereas R4 at 1k trained for two epochs (1,792
seen). At matched examples seen (Figure~\ref{fig:seen}a, $\sim$1.8--2k), R4 (0.296--0.309) lies between R3 (0.267) and
R2 (0.349), so multi-task data does not improve level control per training example. Per \emph{paraphrase} example it is
more economical: R4 at 1k saw about 660 paraphrases, a third of R2 and R3 at 2k, for a level gap in the same range
(Figure~\ref{fig:seen}b).

\paragraph{A small model of VWEN already matches R5 on level control, but not on everything.} R4 at 1k (888 samples,
1,792 seen, 12 minutes on a T4) does not differ from R5\at{1.5k} (24,000 seen, 75 minutes) on dev level gap ($-$0.006
[$-$0.037, 0.024]) or Para-H success (+0.044 [$-$0.024, 0.111]), also on non-questions, and handles the other tasks
(Equiv 0.888, Multi-Para count accuracy 0.996, Gist 0.87). It is clearly worse on Para-L success ($-$0.097 [$-$0.156,
$-$0.039]).

\paragraph{Ceilings differ.} R4's level gap grows slowly (0.296 $\rightarrow$ 0.352 over eight times the data), R2
flattens after 2k and R1 after 4k, while R3 is still rising at 8k (0.359 $\rightarrow$ 0.418). We did not train beyond
8k.

\begin{table*}[t]
\centering\small
\begin{tabular}{@{}llrrrrr@{}}
\toprule
Metric & Recipe & 1k & 2k & 4k & 8k & R5\at{1.5k} (3 seeds) \\
\midrule
\multirow{2}{*}{Para-H meaning (1--5)} & R1--R3 & 4.29--4.57 & 4.39--4.67 & 4.36--4.49 & 4.47--4.62 & \multirow{2}{*}{4.36--4.51} \\
 & R4 & 4.45 & 4.58 & 4.55 & 4.48 & \\
\midrule
\multirow{2}{*}{Simplify rating} & R1--R3 & 0.57--0.73 & 0.67--0.74 & 0.73--0.84 & 0.65--0.77 & \multirow{2}{*}{0.75--0.82} \\
 & R4 & 0.74 & \best{0.83} & 0.81 & 0.81 & \\
\midrule
\multirow{2}{*}{Gist rating} & R1--R3 & 0.47--0.57 & 0.50--0.63 & 0.54--0.56 & 0.48--0.58 & \multirow{2}{*}{0.60--0.64} \\
 & R4 & \best{0.87} & \best{0.91} & \best{0.88} & \best{0.86} & \\
\midrule
\multirow{2}{*}{Style rating} & R1--R3 & 0.44--0.53 & 0.47--0.49 & 0.50--0.52 & 0.45--0.55 & \multirow{2}{*}{0.45--0.54} \\
 & R4 & \best{0.64} & \best{0.64} & \best{0.68} & \best{0.63} & \\
\bottomrule
\end{tabular}
\caption{Dev curves of the LLM-rated metrics (seed 42, 100 outputs per task; intervals roughly $\pm$0.2 on the 1--5
meaning scale and $\pm$0.07 on ratings). R1--R3 are given as a range.}
\label{tab:llm-curves}
\end{table*}

LLM-rated meaning on Para-H depends neither on size nor on recipe (Table~\ref{tab:llm-curves}). R4 is flat on Gist and
Style from 1k (73 Gist and 106 Style samples) and reaches its Simplify plateau at 2k (219 samples), subject to the grader
caveat in Section~\ref{sec:validity}.

\subsection{Held-out Instructions}
\label{sec:heldout}

\begin{table*}[t]
\centering\small
\scalebox{0.95}{
\begin{tabular}{@{}lrrrrrrr@{}}
\toprule
Metric & Base & R1 & R2 & R3 & R4 & R5\at{1.5k} & R5\at{4k} \\
\midrule
Para-H success & 0.08 $\rightarrow$ 0.17 & 0.54 $\rightarrow$ 0.53 & 0.71 $\rightarrow$ 0.66 & 0.64 $\rightarrow$ 0.62 & 0.61 $\rightarrow$ 0.58 & 0.46 $\rightarrow$ 0.44 & 0.53 $\rightarrow$ 0.51 \\
Para-L success & 0.90 $\rightarrow$ 0.87 & 0.66 $\rightarrow$ 0.66 & 0.72 $\rightarrow$ 0.70 & 0.81 $\rightarrow$ 0.78 & 0.73 $\rightarrow$ 0.67 & 0.75 $\rightarrow$ 0.69 & 0.76 $\rightarrow$ 0.68 \\
Equiv accuracy & 0.75 $\rightarrow$ 0.66 & 0.16 $\rightarrow$ 0.23 & 0.14 $\rightarrow$ 0.20 & 0.16 $\rightarrow$ 0.15 & 0.93 $\rightarrow$ 0.89 & 0.16 $\rightarrow$ 0.23 & 0.17 $\rightarrow$ 0.15 \\
Multi-Para count acc. & 0.52 $\rightarrow$ 0.56 & 0 $\rightarrow$ 0 & 0 $\rightarrow$ 0 & 0 $\rightarrow$ 0 & 1.00 $\rightarrow$ 1.00 & 0 $\rightarrow$ 0 & 0 $\rightarrow$ 0 \\
\bottomrule
\end{tabular}
}
\caption{Test scores with the canonical instruction $\rightarrow$ with a held-out phrasing never seen in training (R4 and
R5\at{1.5k}: mean over three seeds).}
\label{tab:heldout}
\end{table*}

Unseen phrasings cost every fine-tuned model 0--0.06 and leave the ranking of recipes unchanged
(Table~\ref{tab:heldout}), so the models respond to the meaning of the instruction rather than to a memorised template.
R4 loses most on Para-L success ($-$0.06), like both R5 runs.

\subsection{Paraphrase Judgement}
\label{sec:equiv}

\begin{figure*}[t]
\centering
\includegraphics[width=\textwidth]{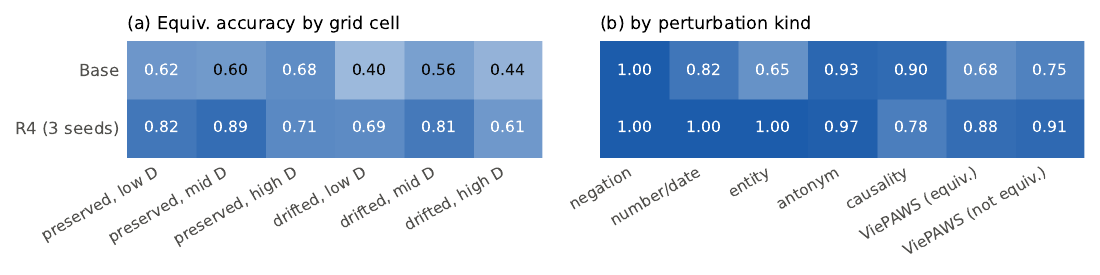}
\caption{Equiv accuracy on test by grid cell (meaning $\times$ divergence) and by perturbation kind. R4: mean over three
seeds.}
\label{fig:judge}
\end{figure*}

Only the base model and R4 answer Equiv with a label. R4 improves on the base model in every grid cell
(Figure~\ref{fig:judge}); the hardest cells are those where surface similarity misleads: drifted pairs with high
divergence (0.61) and preserved pairs with high divergence (0.71). On targeted perturbations R4 reaches 1.00 on negation,
numbers and dates, and entities, and 0.97 on antonyms, but drops below the base model on causality (0.78 vs.\ 0.90).

\section{Error Analysis}
\label{sec:errors}

\paragraph{Answering instead of rephrasing.} ViSP contains almost no questions (0.2\% of inputs). Asked to paraphrase a
question, the R5 models often return a declarative sentence that asserts an answer (Table~\ref{tab:q2s},
Table~\ref{tab:random-examples} example 1). This happens for 16--26\% of question inputs across R5 runs and 23\% for the
base model, but under 1\% for every recipe trained on our pool, which contains question paraphrases from ViQP and
VietQuill.

\begin{table*}[t]
\centering\small
\begin{tabular}{@{}lrrrrrrr@{}}
\toprule
 & Base & R1 & R2 & R3 & R4 & R5\at{1.5k} & R5\at{4k} \\
\midrule
Question $\rightarrow$ statement & 22.6\% & 0.4\% & 0.0\% & 0.2\% & 0.2--0.9\% & 16.4--26.2\% & 21.9\% \\
\ldots\ of which pass MP & 38\% & -- & -- & -- & -- & 60--64\% & 68\% \\
\bottomrule
\end{tabular}
\caption{Share of question inputs (562 Para-L and Para-H requests on test) whose output is not a question, and the share
of those outputs that still pass the MP check. R4, R5\at{1.5k}: three seeds.}
\label{tab:q2s}
\end{table*}

\paragraph{The MP check does not catch it.} Around 60--68\% of these statement outputs from R5 pass bidirectional
entailment: the NLI model treats ``When a person holds the monarchy, they are restricted in power'' and the question it
came from as mutual entailments. MP is therefore an upper bound on true meaning preservation for question inputs, and
R5's measured deficit on questions is if anything understated. A question-form check would be a cheap addition to the
metric.

\begin{table*}[t]
\centering\small
\begin{tabular}{@{}lp{0.68\textwidth}rcc@{}}
\toprule
Model & Output (gloss) & \Dv & MP & Success \\
\midrule
\multicolumn{5}{@{}p{0.95\textwidth}@{}}{\textbf{1.} \vi{Khi một người đảm nhiệm chức vụ quân chủ thì họ sẽ bị hạn chế gì?} \textit{(When a person holds the office of monarch, what restrictions do they face?)}} \\
Base & \vi{Khi một vị trí quân chủ được giữ, người đó sẽ bị hạn chế về quyền lực, quyền tự do và quyền lực chính trị.} \textit{(When a monarchy is held, the person is restricted in power, freedom and political power.)} & 0.76 & \xmark & \xmark \\
R4 & \vi{Những hạn chế nào sẽ đối mặt với người đảm nhiệm chức vụ quân chủ?} \textit{(Which restrictions will face the person holding the office of monarch?)} & 0.85 & \cmark & \cmark \\
R5 & \vi{Khi một người đảm nhiệm chức vụ quân chủ, họ sẽ bị hạn chế quyền lực.} \textit{(When a person holds the office of monarch, they are restricted in power.)} & 0.02 & \cmark & \xmark \\
\midrule
\multicolumn{5}{@{}p{0.95\textwidth}@{}}{\textbf{2.} \vi{Kim cương được khai thác tại đâu?} \textit{(Where are diamonds mined?)}} \\
Base & \vi{Kim cương được khai thác tại đâu?} \textit{(copy of the input)} & 0.00 & \cmark & \xmark \\
R4 & \vi{Địa điểm khai thác kim cương là nơi nào?} \textit{(Which place is the location where diamonds are mined?)} & 0.69 & \cmark & \cmark \\
R5 & \vi{Kim cương được tìm thấy ở đâu?} \textit{(Where are diamonds found?)} & 0.08 & \cmark & \xmark \\
\midrule
\multicolumn{5}{@{}p{0.95\textwidth}@{}}{\textbf{3.} \vi{Lực lượng chủ chốt cho cuộc chiến đấu tranh chống Pháp là đoàn quân đoàn nào?} \textit{(Which army corps was the key force in the struggle against the French?)}} \\
Base & \vi{Đoàn quân đóng vai trò trung tâm trong cuộc chiến tranh chống Pháp, và câu hỏi này đặt ra một vấn đề cốt lõi: ``Đoàn quân nào'' là chính lực chính yếu?} \textit{(The army played a central role in the war against the French, and this question raises a core issue: ``which army'' is the main force?)} & 0.84 & \cmark & \cmark \\
R4 & \vi{Đoàn quân nào là lực lượng chủ chốt trong cuộc chiến đấu chống Pháp?} \textit{(Which army corps was the key force in the fight against the French?)} & 0.63 & \cmark & \cmark \\
R5 & \vi{Đoàn quân nào là lực lượng chủ chốt trong cuộc chiến tranh chống Pháp?} \textit{(Which army corps was the key force in the war against the French?)} & 0.51 & \cmark & \cmark \\
\midrule
\multicolumn{5}{@{}p{0.95\textwidth}@{}}{\textbf{4.} \vi{Naples và Algiers được ngăn cách bởi biển nào?} \textit{(Which sea separates Naples and Algiers?)}} \\
Base & \vi{Naples và Algiers được chia sẻ bởi một biển duy nhất, là Biển Đỏ.} \textit{(Naples and Algiers share a single sea, the Red Sea.)} & 0.22 & \xmark & \xmark \\
R4 & \vi{Biển nào là ranh giới giữa Naples và Algiers?} \textit{(Which sea is the boundary between Naples and Algiers?)} & 0.81 & \cmark & \cmark \\
R5 & \vi{Biển nào ngăn cách Naples và Algiers?} \textit{(Which sea separates Naples and Algiers?)} & 0.63 & \cmark & \cmark \\
\bottomrule
\end{tabular}
\caption{Para-H outputs for four test inputs drawn at random (seed 7), with English glosses. \Dv, MP and success as
scored by the harness. R4 and R5\at{1.5k} are seed 42.}
\label{tab:random-examples}
\end{table*}

Appendix~\ref{app:examples} (Tables~\ref{tab:example-1}--\ref{tab:example-4}) compares all recipes on four further random
inputs, each requested at both levels. At the low level the fine-tuned models produce similar, lightly edited outputs. At
the high level R2 and R3 tend to restructure the sentence (fronting the cause or the question word), while R5\at{1.5k}
keeps the original order on input 1. Input 4 shows the question failure described above in both R5 runs: they answer the
question (``\ldots because the body needs more water''), and the R5\at{4k} answer still passes MP and counts as a
success. The examples also show that no recipe is free of meaning errors at the high level: R1 and R4 both reverse the
question on input 4 (``doesn't help the body retain water'', ``what prevents water loss''), which MP catches, and on
input 2 R3 returns the same sentence for both levels.

The examples in Table~\ref{tab:random-examples} show three more patterns. R5 tends to stay close to the input when asked
for high divergence (examples 1 and 2, $\Dv \leq 0.08$), which matches its lower Para-H \Dv\ (0.47 vs.\ 0.55 for R4).
The base model either copies (example 2) or answers, sometimes with a false answer (example 4, the Red Sea). And success
can reward a verbose, unnatural output (example 3, base model), since \Dv\ and MP do not measure fluency.

\section{Evaluation Validity}
\label{sec:validity}

\paragraph{Input distribution.} Half of the paraphrase inputs are questions, which ViSP lacks. We report the split
(Table~\ref{tab:paired}) and base the main claims on effects that hold on non-questions.

\paragraph{Evaluation inputs pass through the same pipeline.} Dev and test paraphrase references are drawn after stage 3,
so they satisfy the meaning filter and level bands used to build R2--R4. The model rewrites the input, not the reference,
and MP is computed against the input with an NLI model other than the filter; still, success rewards outputs that match
our band definition, which R2--R4 were trained on and R1 and R5 were not (they were levelled at the midpoint without a
middle band).

\paragraph{Metric blind spots.} MP accepts question-to-statement conversions (Section~\ref{sec:errors}). The low band
includes near-copies, which is why the base model scores 0.904 on Para-L success; Para-L success should be read together
with the level gap. \Dv\ and MP ignore fluency (Table~\ref{tab:random-examples}, example 3).

\paragraph{The LLM grader shares a model with a teacher.} Gemma 4 26B is one of the distillation teachers, the stage-4
quality grader, and the rater of Simplify, Gist and Style. R4 is trained on its outputs and may be favoured on these three
tasks. The paraphrase conclusions rely on NLI and \Dv, not on this grader.

\paragraph{Test review is partial.} The full dev set was reviewed by the author; 558 of 2,000 test samples were checked
by hand.

\section{Discussion}
\label{sec:discussion}

\paragraph{Why does the meaning filter help level control?} Unfiltered pairs at high divergence are often not
paraphrases. A model trained on them learns that ``change a lot'' may change meaning; the data support this reading, since
R1's Para-H outputs are both less divergent (\Dv\ 0.516 vs.\ 0.623) and less often meaning-preserving (MP 0.856 vs.\
0.892) than R2's. Removing drifting pairs makes high divergence a property of form alone, which appears easier to learn.

\paragraph{What do the other tasks contribute?} Per training example they do not improve level control
(Section~\ref{sec:q3}); per paraphrase example they appear to substitute for part of the paraphrase data. Simplify, Gist
and Style all request a controlled change, which may teach the model to condition the amount of change on the
instruction; we have not tested this with a leave-one-task-out ablation.

\paragraph{Practical guidance.} About 1k selected multi-task samples give a model that matches a ViSP-trained model on
level control and high-divergence paraphrasing, keeps the base model's other abilities, and trains in minutes; it is
weaker on low-divergence paraphrasing. If only controllable paraphrasing is needed, filtered paraphrase-only data scaled
to 8k gives the strongest level control. Whatever the recipe, the training pool should contain the input types the model
will see: question paraphrases are what ViSP misses.

\section{Conclusion}
\label{sec:conclusion}

VWEN is a 0.8B Vietnamese rewriting model trained on about 7k selected samples. At matched compute it paraphrases with
high divergence more successfully than a model trained on ViSP, on both questions and other inputs; a ViSP model with
four times more examples matches it on non-questions but still answers questions instead of rephrasing them. The
bidirectional meaning filter is the most effective step of the pipeline, and multi-task data keeps abilities that
paraphrase-only training erases. Bidirectional entailment, the standard meaning check, misses outputs that answer a
question, which suggests adding a form check to paraphrase evaluation.

\section*{Limitations}

\begin{itemize}
\item \textbf{One model and one size.} All results are on Qwen3.5-0.8B; IFD is computed with the model being trained, so the selected set may suit Qwen3.5 specifically.
\item \textbf{Seeds.} Only R4 at 8k and R5\at{1.5k} have three seeds; the paired tests for single-seed recipes do not
include training variance.
\item \textbf{Curves on dev only}, single seed; the 1k--4k runs were not evaluated on test.
\item \textbf{Early stopping on a flat validation loss} selected very early checkpoints for paraphrase-only runs at 1k;
we report examples seen, but a fixed-step comparison would be cleaner.
\item \textbf{Question detection} uses a final question mark, which misses questions without one.
\item \textbf{R4 is under budget} (7,087 vs.\ $\sim$8,000 samples).
\item \textbf{No human evaluation} comparing outputs across recipes; no external baselines (an encoder--decoder such as
ViT5, \citealp{phan-etal-2022-vit5}, or a prompted large LLM); the pipeline was not applied to a data source other than
the current pool.
\item \textbf{\Dv\ depends on one parser} (Stanza) and on percentile references computed on our own data.
\end{itemize}

\section*{Ethics Statement}

\paragraph{Data and licences.} We reuse public Vietnamese datasets under their published terms and distil from
open-weight models through a commercial API; per-source licences and the teachers' terms of use will be listed in the
data card. Source sentences include news, administrative and health text and may contain names of
public figures; we did not collect private data.

\paragraph{Use.} A paraphrasing model can be used to disguise plagiarism or to produce many variants of misleading
content. VWEN is small and its outputs remain close to the input's content by design, but it offers no protection against
such uses. The judgement task is a research component and should not be used as a sole check of factual consistency (it
fails on causality, Section~\ref{sec:equiv}).

\paragraph{Compute.} All training used a single T4 GPU; the 22 reported training runs took about 17 GPU hours of
training in total, plus evaluation.

\bibliography{vwen}

\appendix

\section{Pipeline Counts by Task}
\label{app:counts}

\begin{table}[h]
\centering\small
\begin{tabular}{@{}lrrr@{}}
\toprule
Task & Stage 1 in & Stage 1 kept & Stage 2 kept \\
\midrule
Paraphrase & 307,663 & 264,800 & 178,034 \\
Equiv & 17,243 & 16,693 & 16,629 \\
NLI & 5,203 & 4,910 & 4,910 \\
Style & 3,690 & 3,296 & 1,913 \\
Simplify & 3,665 & 3,138 & 1,769 \\
Gist & 2,159 & 1,598 & 1,434 \\
Multi-Para & 1,527 & 1,509 & 1,124 \\
\bottomrule
\end{tabular}
\caption{Samples in and kept at stages 1 and 2 (training split). ``Paraphrase'' counts unlevelled pairs, which stage 3
splits into Para-L and Para-H.}
\label{tab:counts}
\end{table}

\begin{table}[h]
\centering\small
\begin{tabular}{@{}lrrc@{}}
\toprule
Task & Samples & Clusters & IFD p10 / p50 / p90 \\
\midrule
Equiv & 1,481 & 184 & 0.20 / 0.23 / 0.26 \\
Para-L & 1,333 & 194 & 0.18 / 0.24 / 0.30 \\
Para-H & 1,333 & 195 & 0.33 / 0.39 / 0.46 \\
Simplify & 889 & 92 & 0.22 / 0.31 / 0.40 \\
Style & 889 & 113 & 0.26 / 0.39 / 0.52 \\
Multi-Para & 691 & 114 & 0.41 / 0.54 / 0.65 \\
Gist & 593 & 91 & 0.19 / 0.29 / 0.41 \\
\bottomrule
\end{tabular}
\caption{R4 at 8k after stage 4 (before the quality grader): samples, clusters covered, IFD percentiles.}
\label{tab:r4-mix}
\end{table}

\section{Training Details}
\label{app:training}

\begin{table}[h]
\centering\small
\begin{tabular}{@{}lp{0.62\columnwidth}@{}}
\toprule
Item & Value \\
\midrule
Model & Qwen3.5-0.8B, full fine-tuning, float32 weights with fp16 mixed precision (no bf16 on T4) \\
Optimiser & paged 8-bit AdamW, learning rate 1e-5, cosine decay, 5\% warmup, no weight decay, gradient clipping 1.0 \\
Batch and length & $8 \times 2$ accumulation $= 16$; at most 512 tokens \\
Stopping & at most 3 epochs; validation twice per epoch (at most every 250 steps), patience 2; R5 fixed steps \\
Wall time (one T4) & R4 at 8k 71--84 min; R5\at{1.5k} 75 min; R5\at{4k} 193 min; the ten 1k--4k runs 142 min in total \\
Scoring & xlm-roberta-large-xnli (MP), xlm-roberta-large (BERTScore floor), Stanza (\Dv), Gemma 4 26B grader (about USD 0.02 per run) \\
\bottomrule
\end{tabular}
\caption{Hyperparameters and cost.}
\label{tab:hparams}
\end{table}

\begin{table}[h]
\centering\small
\begin{tabular}{@{}lrrrr@{}}
\toprule
Recipe & 1k & 2k & 4k & 8k \\
\midrule
R1 & 496 & 992 & 4,000 & 16,000 \\
R2 & 496 & 1,952 & 3,904 & 7,808 \\
R3 & 496 & 1,952 & 3,936 & 15,680 \\
R4 & 1,792 & 1,760 & 7,104 & 14,144 \\
\midrule
R5 & \multicolumn{4}{c}{24,000 (@1.5k) $\cdot$ 56,000 (@4k)} \\
\bottomrule
\end{tabular}
\caption{Examples seen by the selected checkpoint (best step $\times$ 16), seed 42.}
\label{tab:seen}
\end{table}

\section{Example Outputs at Both Levels}
\label{app:examples}

Tables~\ref{tab:example-1}--\ref{tab:example-4} show the outputs of every model for four test inputs drawn at random
(seed 7; two statements, two questions), each requested at both levels. Syllables that do not occur in the input are
\nw{highlighted}. Each output is followed by its \Dv, MP and success as scored by the harness.

\input{vwen/examples}

\end{document}

%% file: vwen/examples.tex

\begin{table*}[p]
\centering\small
\textbf{Input 1} (statement): \vi{Whedon sau đó giải thích sự xuất hiện của Idris Elba và Hayley Atwell vì sự thâm nhập vào linh hồn của Scarlet Witch.}\\ \textit{(Whedon then explained the appearance of Idris Elba and Hayley Atwell by the intrusion into Scarlet Witch's soul.)}\par\medskip
\begin{tabular}{@{}p{0.13\textwidth}p{0.42\textwidth}p{0.42\textwidth}@{}}
\toprule
Model & Low level requested & High level requested \\
\midrule
Base & \vi{Sau đó, Whedon giải thích sự xuất hiện của Idris Elba và Hayley Atwell \nw{do} \nw{họ} thâm nhập vào linh hồn của Scarlet Witch.}\newline {\scriptsize\color{gray}$D$ 0.07 \textperiodcentered{} MP \cmark \textperiodcentered{} success \cmark} & \vi{Sau đó, \nw{W}.\nw{H}. \nw{Gardner} \nw{đã} \nw{mô} \nw{tả} sự hiện \nw{diện} của Idris Elba và Hayley Atwell \nw{thông} \nw{qua} \nw{việc} thâm nhập vào linh hồn của Scarlet Witch.}\newline {\scriptsize\color{gray}$D$ 0.12 \textperiodcentered{} MP \xmark \textperiodcentered{} success \xmark} \\[3pt]
R1 Random & \vi{Sau đó, Whedon giải thích \nw{cách} Idris Elba và Hayley Atwell xuất hiện vì \nw{họ} \nw{đã} thâm nhập vào linh hồn của Scarlet Witch.}\newline {\scriptsize\color{gray}$D$ 0.22 \textperiodcentered{} MP \cmark \textperiodcentered{} success \xmark} & \vi{Sự xuất hiện của Idris Elba và Hayley Atwell \nw{là} \nw{do} sự thâm nhập vào linh hồn của Scarlet Witch, \nw{theo} \nw{lời} Whedon.}\newline {\scriptsize\color{gray}$D$ 0.21 \textperiodcentered{} MP \cmark \textperiodcentered{} success \xmark} \\[3pt]
R2 +Filter & \vi{Sau đó, Whedon giải thích sự xuất hiện của Idris Elba và Hayley Atwell, \nw{nguyên} \nw{nhân} \nw{là} \nw{do} sự thâm nhập vào linh hồn của Scarlet Witch.}\newline {\scriptsize\color{gray}$D$ 0.06 \textperiodcentered{} MP \cmark \textperiodcentered{} success \cmark} & \vi{Sự thâm nhập vào linh hồn của Scarlet Witch \nw{đã} \nw{dẫn} \nw{đến} sự xuất hiện của Idris Elba và Hayley Atwell, \nw{theo} \nw{cách} giải thích của Whedon.}\newline {\scriptsize\color{gray}$D$ 0.44 \textperiodcentered{} MP \cmark \textperiodcentered{} success \cmark} \\[3pt]
R3 +Coverage & \vi{Sau đó, Whedon giải thích \nw{rằng} sự xuất hiện của Idris Elba và Hayley Atwell \nw{là} \nw{do} \nw{họ} \nw{xâm} nhập vào linh hồn của Scarlet Witch.}\newline {\scriptsize\color{gray}$D$ 0.06 \textperiodcentered{} MP \cmark \textperiodcentered{} success \cmark} & \vi{Sự thâm nhập vào linh hồn của Scarlet Witch \nw{là} \nw{nguyên} \nw{nhân} \nw{dẫn} \nw{đến} sự xuất hiện của Idris Elba và Hayley Atwell, \nw{theo} \nw{lời} giải thích của Whedon.}\newline {\scriptsize\color{gray}$D$ 0.76 \textperiodcentered{} MP \cmark \textperiodcentered{} success \cmark} \\[3pt]
R4 Multi-task & \vi{Sau đó, Whedon giải thích sự xuất hiện của Idris Elba và Hayley Atwell \nw{là} \nw{do} sự thâm nhập vào linh hồn của Scarlet Witch.}\newline {\scriptsize\color{gray}$D$ 0.01 \textperiodcentered{} MP \cmark \textperiodcentered{} success \cmark} & \vi{Sự xuất hiện của Idris Elba và Hayley Atwell \nw{được} Whedon giải thích \nw{là} \nw{do} \nw{khả} \nw{năng} thâm nhập vào linh hồn của Scarlet Witch.}\newline {\scriptsize\color{gray}$D$ 0.26 \textperiodcentered{} MP \cmark \textperiodcentered{} success \xmark} \\[3pt]
R5 @1.5k & \vi{Sau đó, Whedon giải thích sự xuất hiện của Idris Elba và Hayley Atwell \nw{là} \nw{do} sự thâm nhập vào linh hồn của Scarlet Witch.}\newline {\scriptsize\color{gray}$D$ 0.01 \textperiodcentered{} MP \cmark \textperiodcentered{} success \cmark} & \vi{Sau đó, Whedon giải thích \nw{rằng} Idris Elba và Hayley Atwell \nw{đã} thâm nhập vào linh hồn của Scarlet Witch.}\newline {\scriptsize\color{gray}$D$ 0.15 \textperiodcentered{} MP \cmark \textperiodcentered{} success \xmark} \\[3pt]
R5 @4k & \vi{Sau đó, Whedon giải thích \nw{rằng} Idris Elba và Hayley Atwell xuất hiện vì \nw{họ} \nw{đã} thâm nhập vào linh hồn của Scarlet Witch.}\newline {\scriptsize\color{gray}$D$ 0.22 \textperiodcentered{} MP \cmark \textperiodcentered{} success \xmark} & \vi{Sau đó, Whedon giải thích \nw{rằng} Idris Elba và Hayley Atwell \nw{đã} xuất hiện vì \nw{họ} \nw{đã} thâm nhập vào linh hồn của Scarlet Witch.}\newline {\scriptsize\color{gray}$D$ 0.25 \textperiodcentered{} MP \cmark \textperiodcentered{} success \xmark} \\[3pt]
\bottomrule
\end{tabular}
\caption{Example 1 of four test inputs drawn at random (seed 7), requested at both levels. Highlighted syllables do not occur in the input. R4 and R5 @1.5k are seed 42.}
\label{tab:example-1}
\end{table*}

\begin{table*}[p]
\centering\small
\textbf{Input 2} (statement): \vi{Nam Ossetia đã bị Nga kiểm soát.}\\ \textit{(South Ossetia has been controlled by Russia.)}\par\medskip
\begin{tabular}{@{}p{0.13\textwidth}p{0.42\textwidth}p{0.42\textwidth}@{}}
\toprule
Model & Low level requested & High level requested \\
\midrule
Base & \vi{Nam Ossetia \nw{vẫn} bị Nga kiểm soát.}\newline {\scriptsize\color{gray}$D$ 0.01 \textperiodcentered{} MP \cmark \textperiodcentered{} success \cmark} & \vi{Nam Ossetia \nw{đang} \nw{nằm} \nw{dưới} \nw{sự} kiểm soát \nw{của} Nga.}\newline {\scriptsize\color{gray}$D$ 0.25 \textperiodcentered{} MP \cmark \textperiodcentered{} success \xmark} \\[3pt]
R1 Random & \vi{Nam Ossetia \nw{hiện} \nw{đang} \nw{thuộc} \nw{quyền} kiểm soát \nw{của} Nga.}\newline {\scriptsize\color{gray}$D$ 0.34 \textperiodcentered{} MP \cmark \textperiodcentered{} success \xmark} & \vi{Nga \nw{là} \nw{quốc} \nw{gia} kiểm soát Nam Ossetia.}\newline {\scriptsize\color{gray}$D$ 0.81 \textperiodcentered{} MP \cmark \textperiodcentered{} success \cmark} \\[3pt]
R2 +Filter & \vi{Nam Ossetia \nw{hiện} \nw{đang} \nw{nằm} \nw{dưới} \nw{sự} kiểm soát \nw{của} Nga.}\newline {\scriptsize\color{gray}$D$ 0.33 \textperiodcentered{} MP \cmark \textperiodcentered{} success \xmark} & \vi{Nga kiểm soát Nam Ossetia.}\newline {\scriptsize\color{gray}$D$ 0.67 \textperiodcentered{} MP \cmark \textperiodcentered{} success \cmark} \\[3pt]
R3 +Coverage & \vi{Nam Ossetia \nw{nằm} \nw{dưới} \nw{sự} kiểm soát \nw{của} Nga.}\newline {\scriptsize\color{gray}$D$ 0.36 \textperiodcentered{} MP \cmark \textperiodcentered{} success \xmark} & \vi{Nam Ossetia \nw{nằm} \nw{dưới} \nw{sự} kiểm soát \nw{của} Nga.}\newline {\scriptsize\color{gray}$D$ 0.36 \textperiodcentered{} MP \cmark \textperiodcentered{} success \xmark} \\[3pt]
R4 Multi-task & \vi{Nam Ossetia \nw{đang} \nw{nằm} \nw{dưới} \nw{sự} kiểm soát \nw{của} Nga.}\newline {\scriptsize\color{gray}$D$ 0.25 \textperiodcentered{} MP \cmark \textperiodcentered{} success \xmark} & \vi{Nga \nw{là} \nw{quốc} \nw{gia} kiểm soát Nam Ossetia.}\newline {\scriptsize\color{gray}$D$ 0.81 \textperiodcentered{} MP \cmark \textperiodcentered{} success \cmark} \\[3pt]
R5 @1.5k & \vi{Nam Ossetia \nw{thuộc} \nw{quyền} kiểm soát \nw{của} Nga.}\newline {\scriptsize\color{gray}$D$ 0.70 \textperiodcentered{} MP \cmark \textperiodcentered{} success \xmark} & \vi{Nga đã kiểm soát Nam Ossetia.}\newline {\scriptsize\color{gray}$D$ 0.44 \textperiodcentered{} MP \cmark \textperiodcentered{} success \cmark} \\[3pt]
R5 @4k & \vi{Nam Ossetia đã bị Nga kiểm soát.}\newline {\scriptsize\color{gray}$D$ 0.00 \textperiodcentered{} MP \cmark \textperiodcentered{} success \cmark} & \vi{Nga đã kiểm soát Nam Ossetia.}\newline {\scriptsize\color{gray}$D$ 0.44 \textperiodcentered{} MP \cmark \textperiodcentered{} success \cmark} \\[3pt]
\bottomrule
\end{tabular}
\caption{Example 2 of four test inputs drawn at random (seed 7), requested at both levels. Highlighted syllables do not occur in the input. R4 and R5 @1.5k are seed 42.}
\label{tab:example-2}
\end{table*}

\begin{table*}[p]
\centering\small
\textbf{Input 3} (question): \vi{Triều đại nhà Tùy đã sụp đổ vào năm nào?}\\ \textit{(In which year did the Sui dynasty collapse?)}\par\medskip
\begin{tabular}{@{}p{0.13\textwidth}p{0.42\textwidth}p{0.42\textwidth}@{}}
\toprule
Model & Low level requested & High level requested \\
\midrule
Base & \vi{Triều đại nhà Tùy đã sụp đổ vào năm nào?}\newline {\scriptsize\color{gray}$D$ 0.00 \textperiodcentered{} MP \cmark \textperiodcentered{} success \cmark} & \vi{Triều đại nhà Tùy đã sụp đổ vào năm nào?}\newline {\scriptsize\color{gray}$D$ 0.00 \textperiodcentered{} MP \cmark \textperiodcentered{} success \xmark} \\[3pt]
R1 Random & \vi{Triều đại nhà Tùy \nw{kết} \nw{thúc} vào năm \nw{bao} \nw{nhiêu}?}\newline {\scriptsize\color{gray}$D$ 0.14 \textperiodcentered{} MP \cmark \textperiodcentered{} success \cmark} & \vi{Năm nào \nw{đánh} \nw{dấu} \nw{sự} \nw{kết} \nw{thúc} \nw{của} triều đại nhà Tùy?}\newline {\scriptsize\color{gray}$D$ 0.81 \textperiodcentered{} MP \cmark \textperiodcentered{} success \cmark} \\[3pt]
R2 +Filter & \vi{Triều đại nhà Tùy \nw{kết} \nw{thúc} vào năm nào?}\newline {\scriptsize\color{gray}$D$ 0.03 \textperiodcentered{} MP \cmark \textperiodcentered{} success \cmark} & \vi{Năm nào \nw{chứng} \nw{kiến} \nw{sự} sụp đổ \nw{của} nhà Tùy?}\newline {\scriptsize\color{gray}$D$ 0.61 \textperiodcentered{} MP \cmark \textperiodcentered{} success \cmark} \\[3pt]
R3 +Coverage & \vi{Triều đại nhà Tùy \nw{kết} \nw{thúc} vào năm nào?}\newline {\scriptsize\color{gray}$D$ 0.03 \textperiodcentered{} MP \cmark \textperiodcentered{} success \cmark} & \vi{Năm nào \nw{đánh} \nw{dấu} \nw{sự} \nw{kết} \nw{thúc} \nw{của} triều đại nhà Tùy?}\newline {\scriptsize\color{gray}$D$ 0.81 \textperiodcentered{} MP \cmark \textperiodcentered{} success \cmark} \\[3pt]
R4 Multi-task & \vi{Triều đại nhà Tùy sụp đổ vào \nw{thời} \nw{điểm} nào?}\newline {\scriptsize\color{gray}$D$ 0.03 \textperiodcentered{} MP \cmark \textperiodcentered{} success \cmark} & \vi{Năm nào triều đại nhà Tùy đã sụp đổ?}\newline {\scriptsize\color{gray}$D$ 0.53 \textperiodcentered{} MP \cmark \textperiodcentered{} success \cmark} \\[3pt]
R5 @1.5k & \vi{Triều đại nhà Tùy đã \nw{kết} \nw{thúc} vào năm nào?}\newline {\scriptsize\color{gray}$D$ 0.01 \textperiodcentered{} MP \cmark \textperiodcentered{} success \cmark} & \vi{Năm nào triều đại nhà Tùy sụp đổ?}\newline {\scriptsize\color{gray}$D$ 0.56 \textperiodcentered{} MP \cmark \textperiodcentered{} success \cmark} \\[3pt]
R5 @4k & \vi{Năm nào nhà Tùy sụp đổ?}\newline {\scriptsize\color{gray}$D$ 0.45 \textperiodcentered{} MP \cmark \textperiodcentered{} success \xmark} & \vi{Năm nào \nw{đánh} \nw{dấu} \nw{sự} sụp đổ \nw{của} triều đại nhà Tùy?}\newline {\scriptsize\color{gray}$D$ 0.68 \textperiodcentered{} MP \cmark \textperiodcentered{} success \cmark} \\[3pt]
\bottomrule
\end{tabular}
\caption{Example 3 of four test inputs drawn at random (seed 7), requested at both levels. Highlighted syllables do not occur in the input. R4 and R5 @1.5k are seed 42.}
\label{tab:example-3}
\end{table*}

\begin{table*}[p]
\centering\small
\textbf{Input 4} (question): \vi{Tại sao tập luyện trong điều kiện nóng ẩm sẽ khiến cơ thể mất nước nhiều hơn?}\\ \textit{(Why does training in hot and humid conditions make the body lose more water?)}\par\medskip
\begin{tabular}{@{}p{0.13\textwidth}p{0.42\textwidth}p{0.42\textwidth}@{}}
\toprule
Model & Low level requested & High level requested \\
\midrule
Base & \vi{Tại sao tập luyện trong điều kiện nóng ẩm khiến cơ thể mất nước nhiều hơn?}\newline {\scriptsize\color{gray}$D$ 0.00 \textperiodcentered{} MP \cmark \textperiodcentered{} success \cmark} & \vi{Tại sao \nw{việc} tập luyện trong \nw{môi} \nw{trường} nóng ẩm \nw{lại} \nw{gây} mất nước \nw{cho} cơ thể nhiều hơn?}\newline {\scriptsize\color{gray}$D$ 0.18 \textperiodcentered{} MP \cmark \textperiodcentered{} success \xmark} \\[3pt]
R1 Random & \vi{Tại sao tập luyện trong điều kiện nóng ẩm khiến cơ thể mất nước nhiều hơn?}\newline {\scriptsize\color{gray}$D$ 0.00 \textperiodcentered{} MP \cmark \textperiodcentered{} success \cmark} & \vi{Tại sao \nw{việc} tập luyện trong \nw{môi} \nw{trường} nóng ẩm \nw{không} \nw{giúp} cơ thể \nw{giữ} nước?}\newline {\scriptsize\color{gray}$D$ 0.14 \textperiodcentered{} MP \xmark \textperiodcentered{} success \xmark} \\[3pt]
R2 +Filter & \vi{Tại sao \nw{việc} tập luyện trong điều kiện nóng ẩm \nw{lại} khiến cơ thể mất nước nhiều hơn?}\newline {\scriptsize\color{gray}$D$ 0.01 \textperiodcentered{} MP \cmark \textperiodcentered{} success \cmark} & \vi{Cơ thể mất nước nhiều hơn \nw{khi} tập luyện trong điều kiện nóng ẩm \nw{vì} điều \nw{gì}?}\newline {\scriptsize\color{gray}$D$ 0.78 \textperiodcentered{} MP \cmark \textperiodcentered{} success \cmark} \\[3pt]
R3 +Coverage & \vi{\nw{Vì} sao tập luyện trong điều kiện nóng ẩm khiến cơ thể mất nước nhiều hơn?}\newline {\scriptsize\color{gray}$D$ 0.01 \textperiodcentered{} MP \cmark \textperiodcentered{} success \cmark} & \vi{Điều \nw{gì} khiến \nw{việc} tập luyện trong điều kiện nóng ẩm \nw{dẫn} \nw{đến} mất nước nhiều hơn?}\newline {\scriptsize\color{gray}$D$ 0.52 \textperiodcentered{} MP \cmark \textperiodcentered{} success \cmark} \\[3pt]
R4 Multi-task & \vi{Tại sao \nw{việc} tập luyện trong điều kiện nóng ẩm sẽ \nw{làm} cơ thể mất nước nhiều hơn?}\newline {\scriptsize\color{gray}$D$ 0.01 \textperiodcentered{} MP \cmark \textperiodcentered{} success \cmark} & \vi{Điều \nw{gì} \nw{cản} \nw{trở} \nw{quá} \nw{trình} mất nước \nw{khi} tập luyện trong điều kiện nóng ẩm?}\newline {\scriptsize\color{gray}$D$ 0.89 \textperiodcentered{} MP \xmark \textperiodcentered{} success \xmark} \\[3pt]
R5 @1.5k & \vi{Tại sao tập luyện trong điều kiện nóng ẩm sẽ khiến cơ thể mất nước nhiều hơn?}\newline {\scriptsize\color{gray}$D$ 0.00 \textperiodcentered{} MP \cmark \textperiodcentered{} success \cmark} & \vi{Tập luyện trong điều kiện nóng ẩm \nw{có} thể \nw{dẫn} \nw{đến} mất nước nhiều hơn \nw{do} cơ thể \nw{cần} nhiều nước hơn.}\newline {\scriptsize\color{gray}$D$ 0.37 \textperiodcentered{} MP \xmark \textperiodcentered{} success \xmark} \\[3pt]
R5 @4k & \vi{Tại sao tập luyện trong điều kiện nóng ẩm sẽ \nw{dẫn} \nw{đến} mất nước nhiều hơn?}\newline {\scriptsize\color{gray}$D$ 0.03 \textperiodcentered{} MP \cmark \textperiodcentered{} success \cmark} & \vi{Nóng ẩm \nw{là} \nw{môi} \nw{trường} \nw{lý} \nw{tưởng} \nw{cho} cơ thể mất nước nhiều hơn, \nw{do} \nw{đó} \nw{việc} tập luyện trong điều kiện \nw{này} \nw{có} thể \nw{dẫn} \nw{đến} \nw{tình} \nw{trạng} mất nước.}\newline {\scriptsize\color{gray}$D$ 0.82 \textperiodcentered{} MP \cmark \textperiodcentered{} success \cmark} \\[3pt]
\bottomrule
\end{tabular}
\caption{Example 4 of four test inputs drawn at random (seed 7), requested at both levels. Highlighted syllables do not occur in the input. R4 and R5 @1.5k are seed 42.}
\label{tab:example-4}
\end{table*}